\RequirePackage{fix-cm}
\documentclass[twocolumn]{svjour3}          
\smartqed  
\usepackage{graphicx}
\usepackage{amsmath}

\begin{document}

\title{
Experience, Imitation and Reflection; Confucius' Conjecture and Machine Learning 
}


\author{Amir Ramezani Dooraki
}




\institute{Amir Ramezani Dooraki \at
	Department of Artificial Intelligence, Faculty of Computer Science \& Information Technology, University of Malaya \\
	\email{amir.ramezani.my@gmail.com}           
}

\date{Received: date / Accepted: date}

\maketitle

\begin{abstract}
Artificial intelligence recently had a great advancements caused by the emergence of new processing power and machine learning methods. Having said that, the learning capability of artificial intelligence is still at its infancy comparing to the learning capability of human and many animals. Many of the current artificial intelligence applications can only operate in a very orchestrated, specific environments with an extensive training set that exactly describes the conditions that will occur during execution time. Having that in mind, and considering the several existing machine learning methods this question rises that 'What are some of the best ways for a machine to learn?' 
\\
Regarding the learning methods of human, Confucius' point of view is that they are by experience, imitation and reflection. This paper tries to explore and discuss regarding these three ways of learning and their implementations in machines by having a look at how they happen in minds.
\end{abstract}

\section{Introduction}
How minds work, or in another word how a human brain thinks, with the goal of implementing it in machines, is a long-term question in artificial intelligence.
In the recent years, Artificial Intelligence algorithms have demonstrated outstanding progress and nowadays can be found in a variety of applications such as autonomous vehicles, computer games and health care automations. Having said that, the learning capability of artificial intelligence is still in its infancy comparing to the learning capability of humans. They can only operate in a very orchestrated, specific environments with an extensive training set that exactly describes the conditions that will occur during execution time. Systems of this kind are limited to the expertise and educated guesses of their human programmer; they lack the ability to learn in, or tune themselves to real-world environments autonomously or to employ learning in novel situations.
\par
In order to create smarter machines, with an abstract similarity to human minds, design and development of complicated computational algorithms inspired by natural intelligence has already started. Among bio-inspired algorithms those which simulated cognitive architectures are crucial where they are "the computational implementation of a cognitive model, and as such, constitute the substrate for all the cognitive functionalities in robots, like perception, attention, action selection, learning, reasoning, etc"
\cite{DynamicLearning-Bellas2014}.
\\As a result of these efforts, there are several successful methods of machine learning such as reinforcement learning, supervised learning and unsupervised learning where each group has its own branches: temporal difference and actor-critic methods for reinforcement learning, artificial neural network and its deep architectures for supervised learning and self-organizing and clustering methods for unsupervised learning. Saying that, this question rises that "What are some of the best ways for a machine to learn?" 
\\
In this paper, I try to first address this question from a philosophical point of view and then I try to find the machine learning counterpart algorithms or in another word the way that a particular learning method is implemented in machines. Further, I try to discuss about the most important of learning method, considering this paper point of view, and the techniques and methods useful for its implementation in machines.
\section{Learning Methodology}
"By three methods we may learn wisdom: First, by reflection, which is the noblest; Second, by imitation, which is easiest; and third by experience, which is the bitterest"; Confucius. 
\\
It is possible to find similar terminologies to each of the third ways of learning wisdom according to Confucius in machine learning field. "Third by experience, which is the bitterest": this is similar to 'the trial and error part of the reinforcement learning framework'. "Second, by imitation, which is easiest", this is similar to 'learning from demonstration or imitation learning in computer science'. Finally, "First, by reflection, which is the noblest", this important way of learning wisdom ends up creating new ideas and shaping new theories as a result of serious thought or consideration considering a person's ideological framework and according to  what that he is learned through the Second and Thirds methods: imitation and experience. 
\par
By looking at the background of machine learning algorithms, one can see that supervised learning, unsupervised learning and reinforcement learning are studied and used by researchers for long time and in several cases, and they are still some of the important categories of learning in machines.
Nonetheless, it is also possible to categorize them as following: 
\begin{itemize}
\item Learning by Experience
\begin{itemize}
\item Reinforcement Learning 
\item Unsupervised Learning 
\end{itemize}
\item Learning by Imitation
\begin{itemize}
\item Learning From Demonstration
\begin{itemize}
\item Pure Classification 
\item Inverse Reinforcement Learning 
\end{itemize}
\item Supervised Learning 
\end{itemize}
\item Learning by Reflection
\begin{itemize}
\item ?
\end{itemize}
\end{itemize}
In the rest of this paper, I first explain briefly about the 'learning by experience' and 'learning by imitation' in machine learning field and later focus on the 'learning by reflection' and try to finds some references for it among machine learning algorithms.
\section{Learning by Experience}
Learning by trial and error (that is experience) is one of the foremost methods of learning specially for the sensorimotor development stage of intelligent creatures where they get master in moving their motors gradually by trying and fixing their errors.
"By observing the development stages of an infant in terms of learning to articulate words, one can see that he first discovers how to control phonation, then focuses on vocal variations of unarticulated sounds, and finally automatically discovers and focuses on babbling with articulated proto-syllables" \cite{EarlyVocal-Frier2014}. 
In the following, it is explained how this method is implemented in machine learning community under different titles.
\subsection{Reinforcement Learning}
"RL algorithms address the problem of how a behaving agent can learn to approximate an optimal behavioral strategy, usually called a policy, while interacting directly with its environment"\cite{IM-Barto2013}.
\\
Reinforcement learning family of algorithm are important in machine learning community for their high capability in decision making by learning from their own experiences. For an agent to be able to learn in an unknown environment, it should be able to autonomously interact with the environment. The framework of reinforcement learning is able to give the agent such an ability by letting it to take actions and measure the value of each action based on the reward it receives. More actions the agent takes, it gets more information about what is the best move in each state of the environment.
\begin{figure}[h]
	\begin{center}
		\includegraphics[width=0.92\columnwidth]{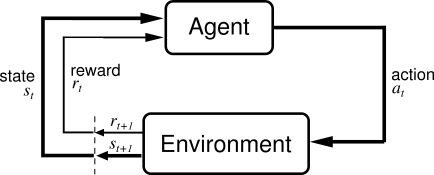}
		\caption{Reinforcement Learning Framework}
		\label{Fig Reinforcement Learning Framework}      
	\end{center}
\end{figure}
\\
Reinforcement learning \cite{1-Sutton1998} algorithms can be divided into several groups such as model-based and model-free, temporal difference methods, policy gradient methods and actor critic methods. Furthermore, they can be based on one or multi-step prediction [5]. An interesting trend in these algorithms is the idea of intrinsic motivation that makes the embodiment of qualities such as curiosity in machines possible. 
\par
The idea of 'learning by experience' in reinforcement learning frame work can be considered as a Markov Decision Process case. Where an agent interacts with its environment at discrete time steps and at each time step agent is located in a unique state according to its sensory input $s_t \in S$. In each $s_t$ agent can take an action, $a_t\in A$ based on its current policy $\pi(s_t)$. In the next time step, the state of agent changes to $s_{(t+1)}\in S$ based on its previous action and a reward R that it receives from the environment. (Figure \ref{Fig Reinforcement Learning Framework}). 
\subsubsection*{Temporal Difference Approaches; Model-based and Model-less}
Temporal difference algorithms are the reinforcement learning algorithms that try to calculate the value of a state or the values of each possible action in that state with attention to the reward received in each state. Furthermore, this information will be used by the agent to use in its policy in which it helps the agent to choose the best action in each state, and as a result the agent finally would be able to maximize its accumulated reward.
\subsubsection*{Q-Learning; Model-less Method}
In many problems, there is no model of the environment. As a result, a model-free off-policy method such as q-learning \cite{14-Watkins1992} can be best to suite the scenario where it can be combined with experience replay method \cite{15-Lin1992} for designing more capable learning algorithms. A temporal difference reinforcement learning method can work for example based on an epsilon-greedy policy for action selection. The goal of this kind of policies is to achieve the maximum discounted reward over time and by using an iterative method of selecting the action with maximum value in each state$-$and updating this value using itself, the maximum action value in the next state and the reward it receives from the environment \cite{1-Sutton1998}. 
\subsubsection*{Dynamic Programing, Monte-Carlo and SARSA; A Model-based Method}
As it can be noticed by looking at the title of a model-based reinforcement learning algorithm, its difference with model-less RL is in having a model of the environment. No matter, how the environment is, either 1D, 2D or 3D environment, or a mathematical formula of the dynamics of the system, a model based algorithm is able to calculate a future state of the environment with attention to the current state and the desired action. A very good example of a model-based algorithm is the work in \cite{Alphago-Silver2016} where a machine is able to play Go game in a human level and win over many human players.
\subsubsection*{Function Approximator}
Many reinforcement learning algorithms use the mentioned formula in order to find the optimal action value. However, there are two main problems, the lack of generalization and the problem of number of the states. In order to tackle these problems a function approximator an be used in order to find the optimal values of each action or state.
As a result of using function approximators, Combination of supervised and unsupervised learning with reinforcement learning can be seen in several works. For example, combination of unsupervised learning (self-organizing map) and reinforcement learning can be seen in several works such as \cite{ContinuousActionQ-Learning-Millan2002} where the input signals of the robot sensors are clustered using a dynamic variation of Kohonen self-organizing map algorithm \cite{SO-Kohonen:2001}, and each cluster is defined to be an state for the reinforcement learning agent. As a result, the RL agent is able to control the robot movement based on pre-defined goals and rewards. 
\\
Further, combination of reinforcement learning algorithm and supervised learning algorithms such as neural network can ends up creating a robust class of algorithm in terms of learning to achieve an optimal behavior similar to a human \cite{3-Mnih2015} in playing Atari game by defining a Deep Q-Network which is as a matter of fact combination q-learning and deep learning \cite{DeepLearning-Hinton2007}. 
\subsubsection*{Actor Critic Approaches}
Based on this view, reinforcement learning methods can be grouped under two categories. Actor-only methods and critic-only methods.
\begin{itemize}
	\item
	"Actor-only methods work with a parameterized family of policies. The gradient of the performance, with respect to the actor policies, is directly estimated by simulation, and the parameters are updated in a direction of improvement." \cite{AC-Konda:2003:AA:942271.942292}
	\item
	"Critic-only methods rely exclusively on value function approximation and aim at learning an approximate solution to the bellman equation, which will the hopefully prescribe a near-optimal policy." \cite{AC-Konda:2003:AA:942271.942292}
\end{itemize}
Actor-critic methods \cite{NeuronlikeBarto1983},
try to mix the important properties of actor-only and critic-only methods. The actor's policy parameters will be updated with the goal of improving the performance using the value function learned by critic through approximation architecture.
"The word actor and critic are synonyms for the policy and action-value function, respectively" \cite{AC-Grondman:2012:SAR:2719653.2719775}.
One of the fresh important works in this area is the deep deterministic policy gradient \cite{DDPG-LillicrapHPHETS2015} where it is an actor-critic network that is consist of two multilayer or deep artificial neural network based actor and critic. This algorithm could successfully solve the problem of continuous action control in some games and classic control problems.
\subsubsection*{Policy Gradient Methods}
As mentioned in the previous section, actor-only methods are those in which the policy is parameterized and can be optimized directly.
"The benefit of a parameterized policy is that a spectrum of continuous actions can be generated, but the optimization methods used (typically called policy gradient methods) suffer from high variance in the estimate of the gradient, leading to slow learning \cite{AC-Konda:2003:AA:942271.942292}\cite{Boyan2002-TechnicalUpdate}\cite{Baxter2001-InfiniteHorizon}\cite{Richter2007-NaturalActorCritic}\cite{Berenji2003-ActorCritic-FRL}." \cite{AC-Grondman:2012:SAR:2719653.2719775}
\subsubsection{Hierarchical Reinforcement Learning Framework}
\label{--- Hierarchical Reinforcement Learning Framework}
Considering human behavior it "has long been recognized to display hierarchical structure: actions fit together into subtasks, which cohere into extended goal-directed activities" \cite{OptBehaviour-ASolway2014}.
\par
"Hierarchical reinforcement learning (HRL) decomposes a reinforcement learning problem into a hierarchy of sub problems or subtasks such that higher-level parent-task invoke lower-level child task as if they were primitive actions. A decomposition may have multiple levels of hierarchy" \cite{HRL-Hengst2010}.
This "decomposition into subproblems has many advantages. First, policies learned in subproblems can be shared (reused) for multiple parent tasks. Second, the value functions learned in subproblems can be shared, so when the subproblem is reused in a new task, learning of the overall value function for the new task is accelerated. Third, if state abstractions can be applied, then the overall value function can be represented compactly as the sum of separate terms that each depends on only a subset of the state variables. This more compact representation of the value function will require less data to learn, and hence, learning will be faster."\cite{HRL-MAXQ-Dietterich2000}
\label{Paper: Hierarchical Reinforcement Learning}
\label{Paper: Optimal Behavioral Hierarchy}
\label{Paper: Hierarchical Reinforcement Learning with the {MAXQ} Value Function Decomposition}
\begin{figure}[h]
	\begin{center}
		\includegraphics[width=0.99\columnwidth]{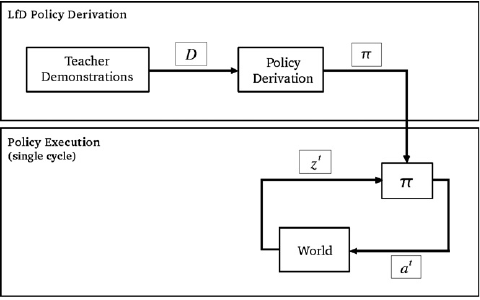}
		\caption{Learning from Demonstration framework, Adapted from \cite{LfDSurvey-Argall2009}}
		\label{Figure: Learning from Demonstration} 
	\end{center}    
\end{figure}
\subsection{Unsupervised Learning}
Unsupervised learning method cannot be categorized under learning that happens by imitation since there is no output for the input to be imitated by machine, and it is hard to categorize it under learning that happens by experience perhaps because it does not have an attached reward to classify that particular experience as good or bad. However, this paper argues that receiving reward happens through many time steps where each time step can be considered as a new experience for our machine, that is each new sample has a subtle reward (loss) that tries to move a particular input toward a particular class. Thus it can be concluded that it is under the category of 'learning by experience'.
\section{Learning by Imitation}
The learning capability of human kind develops throughout his life by getting master in 'learning by imitation' (e.g. learning how to write a word from the first grade teacher), a wide spectrum of what we learn after our early ages comes from this type of learning. 
\subsection{Learning From Demonstration (Apprenticeship Learning)}
\label{--- Learning From Demonstration (Apprenticeship Learning)}
Learning from Demonstration is a natural and intuitive machine learning approach where an agent that called apprentice tries to learn a behavior from demonstration of another agent called the expert. "We simply show the robot how to achieve a task. This has the immediate advantage of requiring no (or very little) specialized skill or training, and makes use of a human demonstrator's existing procedural knowledge useful to identify which control program to acquire" \cite{LfD-Konidaris12a}. 
\label{Robot Learning from Demonstration by Constructing Skill Trees}
A very good example for learning from demonstration, is how to fly an airplane, considering all the different possibilities exist for using the buttons and handles in the cabin of an airplane, the best way to learn how to fly is by observing an expert. Perhaps it is just impossible to try to learn to fly an airplane simply by random pushing of buttons.
Looking at the literatures there are at least two methods to deal with this problem of Apprenticeship Learning (AL); Pure Classification and Inverse Reinforcement Learning (IRL)
\subsection{Pure Classification}
Pure classification, for instance, the large margin method \cite{Ratliff2008-PureClassification} has several advantages and also some disadvantages that is explained in the following; 
It is "easy to implement, fast, no need to resolve MDPs, no need to do a choice of features thanks to boosting techniques. However, they do not take into account the structure of the MDP because the temporal structure of the expert trajectories is not used in a pure classification method. To tackle this drawback, the authors of \cite{Melo2010-LearningFromDemonstation} use a kernel-based approach to encapsulate the structure of the MDP into the classifier, which needs the calculation of the MDP metrics and thus the knowledge of the whole dynamics." \cite{Apprenticeship-Piot:2014:BRC:2615731.2617447}  
\subsection{Inverse Reinforcement Learning}
Inverse Reinforcement Learning (IRL), introduced in \cite{Russell-LearningAgent-1998} and formalized in \cite{Ng-AlgorithmsFor-2000}, is related to Apprenticeship Learning (AL) family of algorithm. However, IRL searches to find a reward where this reward should be able to explain the expert behavior and not the expert policy.
"The key idea behind IRL is that the reward may be the most succinct hypothesis explaining the expert behavior. Some algorithms \cite{Abbeel2004-Apprenticeship}\cite{Neu2009-Training}\cite{Syed2008-AGame} use IRL as an intermediary step to find a policy but other algorithms \cite{Boularias2011-RelativeEntropy}\cite{Klein2012-InverseReinforcement} are "pure" IRL algorithms and output a reward. This reward must then be optimized via a direct reinforcement learning algorithm." \cite{Apprenticeship-Piot:2014:BRC:2615731.2617447}
\subsection{Supervised Learning}
This famous method of machine learning helps the machine to learn to imitate that if a particular input received what should be the particular output. As a result, in this paper it is considered to be under the category of 'learning by imitation'.
\section{Learning by Reflection}
If we describe 'learning by reflection' as the result of serious thought and consideration which can cause for example the creation of  a scientific discovery or theory, then it can be explained as a logical outcome of the following cognitive activities:
\\
\subsection{Reasoning (Deduction and Induction)}
Many of the scientific (e.g computation, biological, social) theories are as a result of a chain of reasonings. Where one can interpret reasoning as an unsupervised or supervised pattern recognition, where for instance by seeing a view a person can recognize it or categorize it as a special event or group, and as a result it can be categorized under the 'learning by experience' or 'learning by imitation'. \\
Reasonings help us to understand and being able to explain where an incident is coming from or where it will ends up or what is the meaning of a particular perception (e.g. a view, a sound, an smell or a combination of several senses (biological sensors)). Furthermore, following the trajectory of incidents (started by incident 'a') using chain of reasonings one can end up to a particular incident 'b' and form a social or scientific theory, where it begins by incident 'a' and ends up to incident 'b'. For example, a person look is faint, we can reason that its because he is hungry or sick, let say the person is hungry thus he does not have money or is very lazy to buy food, or let say the person is sick so he does not have money to buy food or he got a virus, perhaps his immune system is weak or the virus is new, and this chain of reasoning can continue to reach a particular theory based on the look of a person.
\subsection{Imagination}
When a person imagines, he creates or shapes a perception in his mind, which has been obtained according to what he has perceived previously (either through experience or imitation) with attention to his current thoughts and ideas; this imagination can be altered or enhanced with-in the framework of his believes (e.g. a scientific belief or a religious belief).
\\
Here, I emphasize that even though machine imagination concept that is discussed in \cite{MachineImaginationWrong-KatoH2015} and similar works are interesting but they are not the topic that this paper intends to discuss about. In those works imagination for machine is discussed and explained as a process that convert an input, for example a text, to an image where the process and type of learning can be categorized more under the category of 'learning by imitation' (learning to produce a particular image when input is a particular text or sentence). Nonetheless, what this paper means by "imagination" in a machine is as a matter of fact a process that uses machine memory and knowledge in order to create new perceptions that did not shown to machine before and in order to learn from this new perception.
\par
This paper categorizes imagination into the following categories: 
\begin{itemize}
	\item 
	The first category is a synthesization of world model, based on the past experiences, for example, imagine a person that seen a place before, he is able to imagine this place, or world model, again in his mind and take new actions and calculates new results accordingly. In computer science there are instances of this kind of imagination, for instance SLAM \cite{Durrant-Whyte2006}, where a machine learn a model of the world and as a result imagines this model and particular outcomes with-in this imaginary model of the world according to particular actions. Model-based reinforcement learning algorithms are among those in which a model of the world can be built, and then refined and completed using experiences and observations. Model-based RL aims to endow an agent "with a model of the world, synthesized from past experience. By using an internal model to reason about the future, here also referred to as imagining, the agent can seek positive outcomes while avoiding the adverse consequences of trial-and-error in the real environment – including making irreversible, poor decisions"
	\cite{Weber2017arc}.
	\\
	One important consideration in this type of imagination is that it is limited by the model of the world, made by observation, that is, this kind of imagination cannot go beyond the observation of person or machine. In another world, this imagination is limited to experience and observation and it is kind of learning by experience.
	Thus, perhaps it is possible to think that model-based thinking (method) is a reasoning process which of course can reach to very important discoveries and outcomes.
	\item
	In order to explain the second category, first, let's imagine Einstein thinking about time being relative, he is considering that his bus can move faster than a beam of light coming from the clock tower, this idea perhaps is not based on experience or observation (it is more than a model-based method). 
	He was imagining something out of the world-model he experienced by creating at the first point an exception beyond or contradictory to the experiences and observations he had because nobody can move faster than light so it can not be experienced nor it can not be imitated, however, this way of thinking (exceptional or contradictory) paved the way to create a great theory. 
	\\
	Furthermore, following the chain of reasonings it is possible to reach a point that the person does not have any experience or imitation information about it, for example considering biological advancements and discovery of a method that makes human immortal, the fact of immortal human perhaps is not experienced before, it can be imitated from the movies, but for the first time there was no movie about it, this example is also an imagination that is based on pure imagination contradictory or beyond experiences and imitations.
\end{itemize}
Is it possible to implement the 'learning by reflection' in machines? As explained in this section, a reflection can be explained as an imagination that is not limited to experience or imitation. Chain of reasonings and limited imagination both can be explained by 'learning by experience' or 'learning by imitation', however, a reflection, needs an elaboration, enhancement, or exception beyond the experienced world-model. This enhancement would happen according to the believe frame-work of the machine or person where this belief frame work can be scientific, religious or spiritual. Further, this enhancement can happen using what learned by imitation or by mixing different experiences into each other and judge the possibility of the result according to the believe framework of the person or machine.
\section{Conclusion} 
In this paper, several ways of learning for a machine $-$learning by experience, learning by imitation and learning by reflection$-$ inspired by \textit{human ways of learning wisdom explained by Confucius}, introduced and discussed. Further, it explained that two of these methods, 'learning by experience' and 'learning by imitation' are widely researched and implemented in machine learning community under other titles. For example, \textit{reinforcement learning} and \textit{unsupervised learning} are based on 'learning by experience' and \textit{learning from demonstration} and \textit{supervised learning} are based on 'learning by imitation'. Nevertheless, the third way of learning wisdom, 'learning by reflection', is focused less and less implemented as well in machine learning community and algorithms. As a result, this paper discussed about this method of learning and the ways it can be explained through them in minds and machines.



\bibliographystyle{spphys}
\bibliography{references}

\end{document}